\documentclass{INTERSPEECH2023}

\interspeechcameraready

\usepackage{multirow}
\usepackage{acro}
\usepackage{soul}
\usepackage{microtype}
\usepackage[normalem]{ulem}
\DeclareAcronym{davenet}{
	short = DAVEnet ,
	long  = deep audio-visual embedding network,
	alt = Deep Audio-Visual Embedding Network
}
\DeclareAcronym{resdavenet}{
	short = ResDAVEnet,
	long  = residual deep audio-visual embedding network,
	alt = Residual Deep Audio-Visual Embedding network
}
\DeclareAcronym{AE}{
	short = AE ,
	long  = autoencoder
}
\DeclareAcronym{CAE}{
	short = CAE,
	long  = correspondence autoencoder,
	alt = Correspondence Autoencoder
}
\DeclareAcronym{MCAE}{
	short = MCAE,
	long  = multimodal correspondence autoencoder,
	alt = Multimodal Correspondence Autoencoder
}
\DeclareAcronym{MTriplet}{
	short = MTriplet,
	long  = multimodal triplet network,
	alt = Multimodal Triplet Network
}
\DeclareAcronym{FFNN}{
	short = FFNN,
	long  = feedforward neural network,
	alt = Feedforward Neural Networks
}
\DeclareAcronym{CNN}{
	short = CNN,
	long  = convolutional neural network,
	alt = Convolutional Neural Networks
}
\DeclareAcronym{RNN}{
	short = RNN,
	long  = recurrent neural network,
	alt = Recurrent Neural Networks
}
\DeclareAcronym{VQ}{
	short = VQ,
	long  = vector-quantisation
}
\DeclareAcronym{mfcc}{
	short = MFCC,
	long  = mel-frequency cepstral coefficient
}
\DeclareAcronym{CPC}{
	short = CPC,
	long  = contrastive predictive coding,
	alt = Contrastive Predictive Coding
}
\DeclareAcronym{multdavenet}{
	short = MultDAVEnet,
	long  = multilingual deep audio-visual embedding network,
	alt = Multilingual Deep Audio-Visual Embedding Network
}
\DeclareAcronym{accmultdavenet}{
	short = AcstMultDAVEnet,
	long  = acoustic multilingual deep audio-visual embedding network,
	alt = Acoustic Multilingual Deep Audio-Visual Embedding Network
}
\DeclareAcronym{DNN}{
	short = DNN,
	long  = deep neural network
}
\DeclareAcronym{ASR}{
	short = ASR,
	long  = automatic speech recognition
}
\DeclareAcronym{AWE}{
	short = AWE,
	long  = acoustic word embedding,
	alt = Acoustic Word Embedding
}
\DeclareAcronym{AGWE}{
	short = AGWE,
	long  = acoustically grounded word embedding,
	alt = Acoustically Grounded Word Embedding
}
\DeclareAcronym{ASE}{
	short = ASE,
	long  = acoustic span embedding,
	alt = Acoustic Span Embedding
}
\DeclareAcronym{QbE}{
	short = QbE,
	long  = query-by-example,
	alt  = Query-by-Example
}
\DeclareAcronym{STD}{
	short = STD,
	long  = spoken term detection
}
\DeclareAcronym{UTD}{
	short = UTD, 
	long = long short-term memory
}
\DeclareAcronym{SGD}{
	short = SGD, 
	long = stochastic gradient decent,
	alt = Stochastic Gradient Decent
}
\DeclareAcronym{bow}{
	short = BoW, 
	long = bag-of-words,
	alt = Bag-of-Words
}
\DeclareAcronym{mse}{
	short = MSE, 
	long = mean squared error,
	alt = Mean Squared Error
}
\DeclareAcronym{vpkl}{
	short = VPKL, 
	long = visually prompted keyword localisation,
	alt = Visually Prompted Keyword Localisation
}

\newcommand{\model}{{\small \scshape MattNet}}
\newcommand{\smallmodel}{{\scriptsize \scshape MattNet}}
\newcommand{\qbert}{{\small QbERT}}

\usepackage[prependcaption,textsize=scriptsize]{todonotes}
\definecolor{mycolor}{HTML}{008000}

\title{Visually grounded few-shot word acquisition with fewer shots}
\name{Leanne Nortje, Benjamin van Niekerk, Herman Kamper}
\address{
	MediaLab, E\&E Engineering, Stellenbosch University, South Africa}
\email{nortjeleanne@gmail.com, benjamin.l.van.niekerk@gmail.com, kamperh@sun.ac.za}

\usepackage{hyperref}
\hypersetup{
	colorlinks=true,
	linkcolor=blue,
	filecolor=magenta,      
	urlcolor=blue
}
\usepackage{cite}

\begin{document}

\maketitle

\begin{abstract}
	We propose a visually grounded speech model that acquires new words and their visual depictions from just a few word-image example pairs.
	Given a set of test images and a spoken query, we ask the model which image depicts the query word.
	Previous work has simplified this problem by either using an artificial setting with digit word-image pairs or by using a large number of examples per class.
	We propose an approach that can work on natural word-image pairs but with less examples, i.e.\ fewer shots.
	Our approach involves using the given word-image example pairs to mine new unsupervised word-image training pairs from large collections of unlabelled speech and images.
	Additionally, we use a word-to-image attention mechanism to determine word-image similarity. 
	With this new model, we achieve better performance with fewer shots 
	than any existing approach.
\end{abstract}

\section{Introduction}
\label{sec:intro}

Speech recognition for low-resource languages faces a major obstacle: it requires large amounts of transcribed data for development~\cite{besacier_automatic_2014}.
To overcome this, we can look to how children acquire new words from a few examples without the use of transcriptions~\cite{biederman_recognition-by-components:_1987, miller_how_1987, gomez_infant_2000, lake_one-shot_2014, rasanen_joint_2015}.
E.g.\ Borovsky et al.\ \cite{borovsky_once_2012} shows that children can acquire a word for a visual object after seeing it only once.
This has led to recent studies into multimodal few-shot learning~\cite{eloff_multimodal_2019, nortje_unsupervised_2020, nortje_direct_2021}: the task of learning new concepts from a few examples, where each example consists of 
instances of the same concept but from different modalities. 
E.g.\ imagine a robot seeing a picture of a \textit{zebra}, \textit{kite} and \textit{sheep}
while also hearing the spoken word for each concept. 
After seeing this small set of examples (called a support set) 
the robot is prompted to identify which image in an unseen set corresponds to the word ``zebra''.

Building off of a growing number of studies in visually grounded speech modelling~\cite{harwath_jointly_2018, kamper_semantic_2019-1, olaleye_attention-based_2021, chrupala_visually_2022, merkx_modelling_2022}, we consider this multimodal problem of learning the spoken form of a word and its visual depiction from only a few paired word-image examples.
Multimodal few-shot speech-image learning was first introduced in~\cite{eloff_multimodal_2019} and then extended in~\cite{nortje_unsupervised_2020} and~\cite{nortje_direct_2021}.
But these studies were performed in an artificial setting where spoken isolated digits were paired with MNIST images of digits.
This shortcoming was recently addressed by Miller and Harwath~\cite{miller_t_exploring_2022}, who considered multimodal few-shot learning on isolated words paired with natural images.
Their specific focus was on learning a new concept while not forgetting previously learned concepts, i.e.\ dealing with the problem of catastrophic forgetting.  
While their methods performed well in a few-shot retrieval task with five classes, it required a relatively large number of samples per class, i.e. many ``shots''.
Our aim is 
to do visually grounded multimodal few-shot learning on natural images 
with fewer shots.
We do not explicitly focus on the catastrophic forgetting problem (for now), although we do evaluate using the same 
setup as~\cite{miller_t_exploring_2022}. 

There are two core components to our new approach.
Firstly, we use the 
support set to ``mine'' new noisy word-image pairs
from unlabelled speech and image collections.
Concretely, each spoken word example in the support set is compared to each utterance in an unlabelled speech corpus; 
we  use a new query-by-example approach (called QbERT) to identify segments in the search utterances that match the word in the support set.
We follow a similar approach for mining additional images from the few-shot classes by using AlexNet~\cite{krizhevsky_imagenet_2017} embeddings and cosine distance for the comparisons between a support set image and unlabelled search images.
The mined words and images are then paired up, thereby artificially increasing the size of our support set (in an unsupervised way).
This pair mining scheme is very similar to that followed in~\cite{nortje_direct_2021}, where it was used on digit image-speech data with simpler within-modality comparisons.

Secondly, our new approach is based on a model with a word-to-image attention mechanism. 
This multimodal attention network (\model) takes a single word embedding and calculates its correspondence to each pixel embedding to learn how the word is depicted within an image. 
This 
is similar to the vision attention part of the model from~\cite{nortje_towards_2022}, where the goal was to localise visual keywords in speech 
(not in a few-shot setting).

We first evaluate our approach on the few-shot retrieval task also used in~\cite{miller_t_exploring_2022}.
We show that \model\ 
achieves higher retrieval scores for fewer shots than \cite{miller_t_exploring_2022}'s models.
We also show that our approach yields more consistent scores with a larger number of few-shot classes.
Secondly, we evaluate our approach in a more conventional few-shot classification task where it only needs to correctly distinguish between classes seen in the support set. We consider settings with different numbers of classes and shots, and show that we can achieve accuracies higher than 60\% with as little as five shots.

Our core contributions is the new mining scheme operating on natural images and speech, and then the application of a new attention-based model to the task of multimodal few-shot learning. 
Our experiments show that both these contributions lead to improvements over previous methods.

\section{Visually grounded few-shot learning and evaluation}
\label{sec:task}

\begin{figure}[tb]
	\centering
	\includegraphics[width=\linewidth]{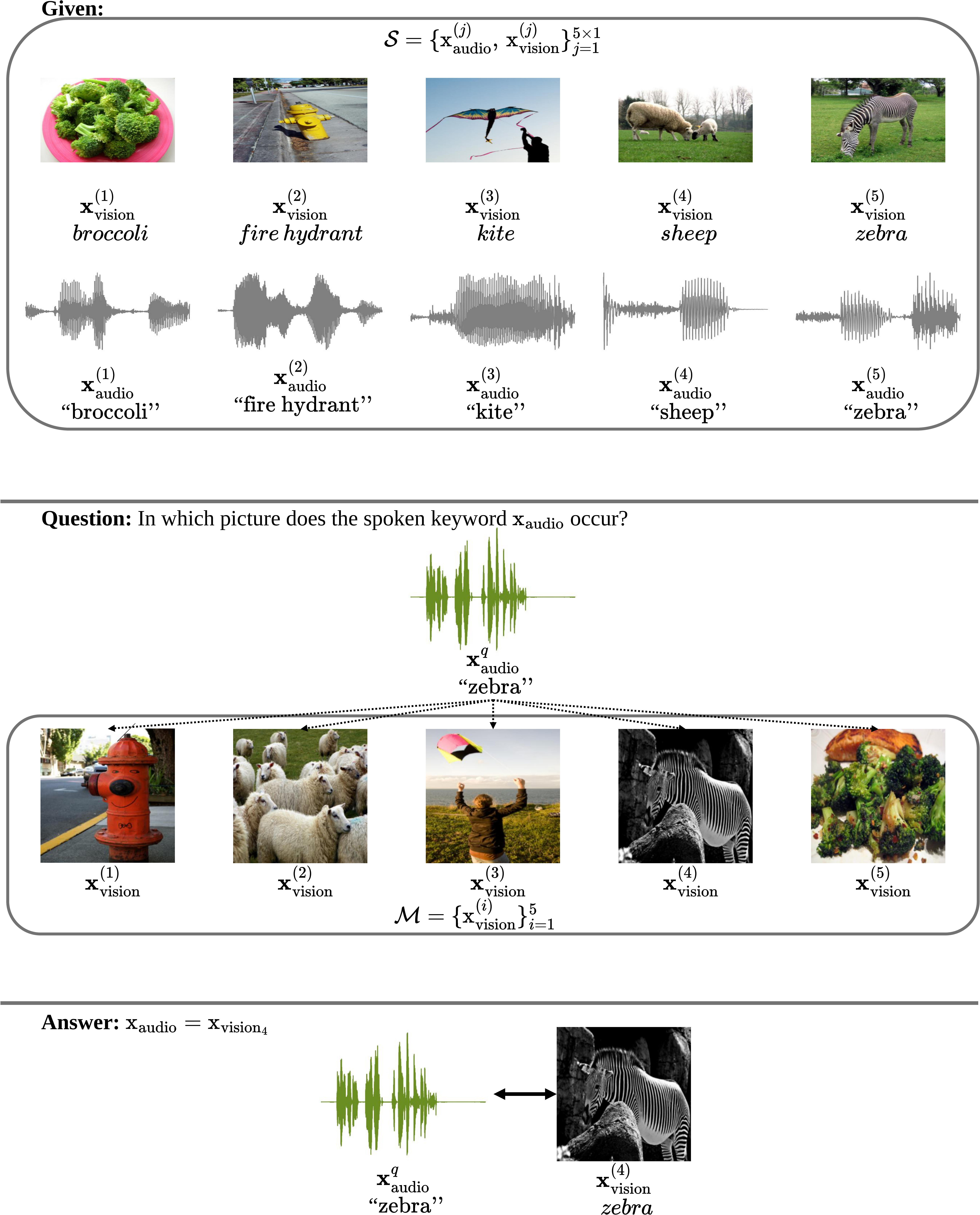}
	\caption{Given the few examples in the support set $\mathcal{S}$, the multimodal few-shot classification task is to e.g.\ identify the image depicting the word ``zebra'' from a set of unseen images.}
	\label{fig:task}
\end{figure}

Below we describe the few-shot learning setup as well as the two tasks that we consider in this work.

\textbf{Visually grounded few-shot learning.} 
Children can learn a new word for a visual object from only a few examples~\cite{borovsky_once_2012}.
To attempt to replicate this in a machine learning model, we train a model on a few spoken word-image examples.
This set of $K$ examples per class is called the support set $\mathcal{S}$.
Each pair in $\mathcal{S}$ consists of an isolated spoken word $\boldsymbol{\mathrm{x}}_{\textrm{audio}}^{(j)}$ and a corresponding image $\boldsymbol{\mathrm{x}}_{\textrm{vision}}^{(j)}$.
For the \textit{one-shot} case shown in the top part of Figure~\ref{fig:task}, $\mathcal{S}$ consists of one word-image example pair for each of the $L$ classes. 
For the $L$-way $K$-shot task, the support set $\mathcal{S}=\{\boldsymbol{\mathrm{x}}_{\mathrm{audio}}^{(j)}, \boldsymbol{\mathrm{x}}_{\mathrm{vision}}^{(j)}\}_{j=1}^{L\times K}$ contains $K$ word-image example pairs for each of the $L$ classes.  


\textbf{Visually grounded few-shot word classification.}
In this task, illustrated in the middle and lower parts of Figure~\ref{fig:task}, we are given an unseen isolated spoken word query $\boldsymbol{\mathrm{x}}^q_{\textrm{audio}}$ and prompted to identify the corresponding image in a matching set $\mathcal{M}=\{(\boldsymbol{\mathrm{x}}_{\textrm{vision}}^{(i)})\}_{i=1}^L$ of unseen test images.
$\mathcal{M}$ contains one image depicting each of the $L$ classes.
Neither the test-time speech query $\boldsymbol{\mathrm{x}}_{\textrm{audio}}$ nor any images in $\mathcal{M}$ are duplicated in the support set.
This image-speech task was considered in~\cite{eloff_multimodal_2019,nortje_unsupervised_2020,nortje_direct_2021}, but here, for the first time, we use natural images instead of isolated digit images.
In contrast to the task that we describe next, this is a conventional few-shot classification task where the model only needs to correctly distinguish between classes seen in the support set, i.e.\ there are no other background or imposter classes.

\textbf{Visually grounded few-shot retrieval.}
In contrast, in this task the goal is to test whether a model 
can search through a large collection of images and retrieve those that depict 
a few-shot query, i.e.\ the matching set 
$\mathcal{M}$ in this case contains images that depict the $L$ few-shot classes but also images that depict other classes.
These additional images might contain completely unseen classes, or background classes potentially seen during pretraining of the few-shot 
model.
The model is penalised if it retrieves one of these imposter images.
This few-shot retrieval task was proposed in~\cite{miller_t_exploring_2022}.  
Their interest was specifically in measuring catastrophic forgetting. 
Since their task requires a model to distinguish between few-shot classes and other classes, it can be used to not only determine whether models can be updated to learn new classes from only a few examples, but also how well the model remembers previously learned (background) classes.
We do not explicitly focus on the catastrophic forgetting problem, but we want to compare to~\cite{miller_t_exploring_2022}, which is why we also consider this retrieval task.

For both tasks we need a distance metric $D_\mathcal{S}(\boldsymbol{\mathrm{x}}^q_{\textrm{audio}},\boldsymbol{\mathrm{x}}_{\textrm{vision}}^{(i)})$ between instances from the speech and vision modalities. 
Below we describe our model that we use as this distance metric.

\section{Multimodal few-shot attention}
\label{sec:model}

Our approach to determine $D_\mathcal{S}(\boldsymbol{\mathrm{x}}^q_{\textrm{audio}},\boldsymbol{\mathrm{x}}_{\textrm{vision}}^{(i)})$ relies on two core components: a model with a word-to-image attention mechanism and a method to mine pairs using a few ground truth word-image examples (given in the support set).

\subsection{Word-to-image attention mechanism}
\label{subsec:model}
\begin{figure}[tb]
	\centering
	\includegraphics[width=0.8\linewidth]{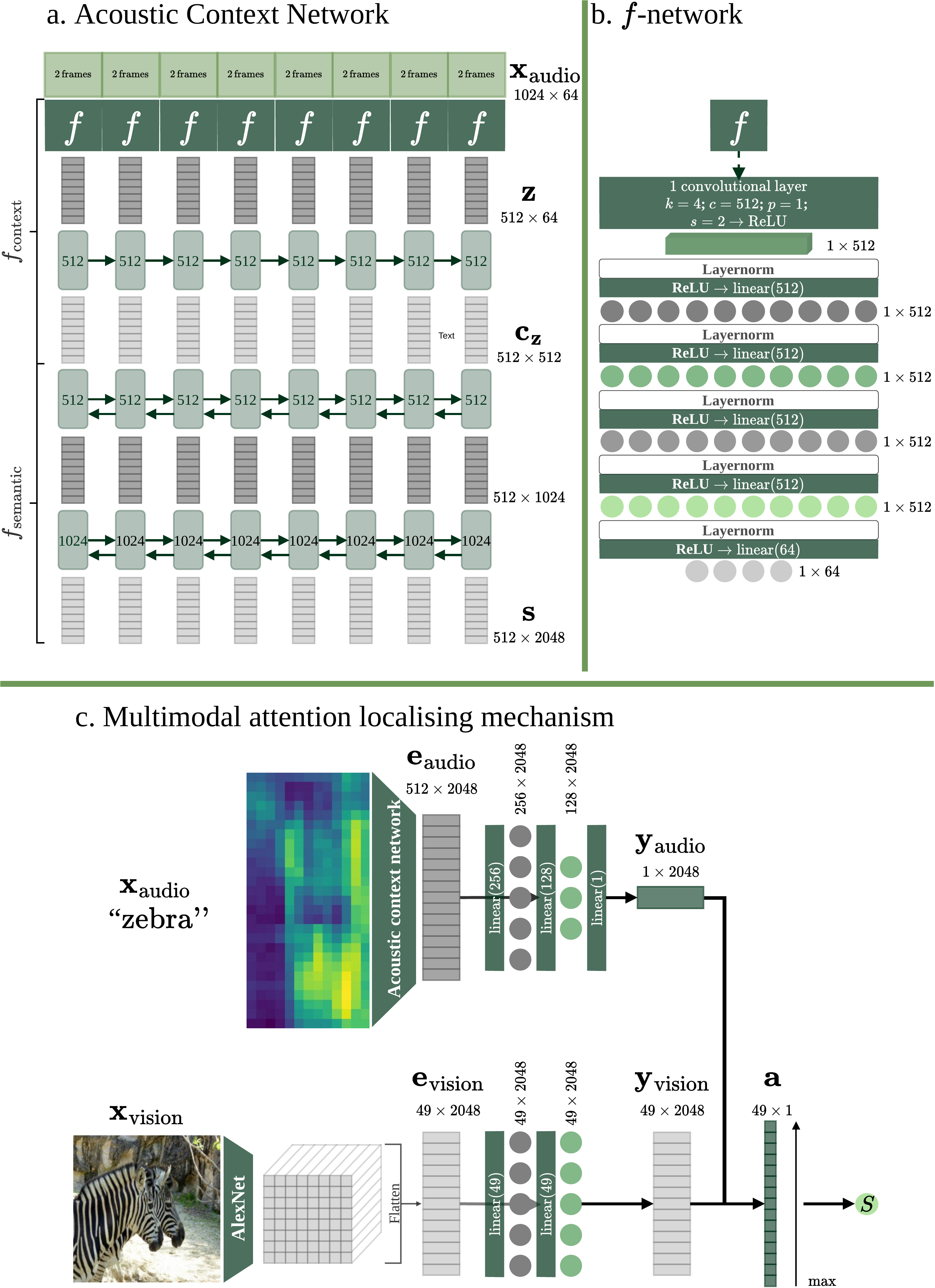}
	\caption{\model\ consists of (c) a vision and an audio network. The audio network consists of (a + b) an acoustic context network and an BiLSTM network. These networks are connected with a word-to-image attention mechanism.}
	\label{fig:model}
\end{figure}

Our model is shown in Figure~\ref{fig:model} and we call it \model\ (\textbf{m}ultimodal \textbf{att}ention \textbf{net}work).
To start off, 
we adapt the multimodal localising attention model of \cite{nortje_towards_2022} that consists of an audio and a vision branch.
For the vision branch, we replace ResNet50~\cite{he_deep_2016} with an adaptation of AlexNet~\cite{krizhevsky_imagenet_2017} to encode an image input $\boldsymbol{\mathrm{x}}_{\textrm{vision}}$ into a sequence of embeddings $\boldsymbol{\mathrm{y}}_{\textrm{vision}}$.
For the audio branch, we use the same audio subnetwork as \cite{nortje_towards_2022} that consists of an acoustic network $f_{\textrm{acoustic}}$ which extracts speech features from a spoken input $\boldsymbol{\mathrm{x}}_{\textrm{audio}}$.
However, 
\cite{nortje_towards_2022} takes
an entire spoken utterance as $\boldsymbol{\mathrm{x}}_{\textrm{audio}}$, whereas we use a single isolated spoken word.
We also add a few linear layers to the BiLSTM network $f_{\textrm{BiLSTM}}$ 
to encode the speech features into a single audio embedding $\boldsymbol{\mathrm{y}}_{\textrm{audio}}$, similar to acoustic word embeddings~\cite{kamper_truly_2019, chung_unsupervised_2016, wang_segmental_2018, holzenberger_learning_2018}.
We connect the vision and audio branches with a multimodal attention mechanism to compare 
the word embedding $\boldsymbol{\mathrm{y}}_{\textrm{audio}}$ to each pixel embedding in $\boldsymbol{\mathrm{y}}_{\textrm{vision}}$.

To get this word-to-image attention mechanism, we take the keyword localising attention mechanism of \cite{nortje_towards_2022} which detects whether certain keywords occur in both spoken utterances and images.
However, we aim to only detect whether a single isolated spoken word occurs somewhere within an image.
More specifically, we calculate attention weights $\boldsymbol{\mathrm{a}}$ over the pixel embeddings by calculating the dot product between $\boldsymbol{\mathrm{y}}_{\textrm{audio}}$ and each pixel embedding in $\boldsymbol{\mathrm{y}}_{\textrm{vision}}$.
By taking the maximum over $\boldsymbol{\mathrm{a}}$, we get a similarity score $S\in[0, 100]$.
The higher $S$, the more probable it is that the spoken word corresponds to one or more objects in the image.
If $S$ is low, it is less probable that any object in the image corresponds to the spoken word.

We train \model\ by using $S$ in a contrastive loss:
{\tiny\begin{equation}\begin{split}		
		l &= \textrm{MSE}\Bigg(S(\boldsymbol{\mathrm{e}}_{\textrm{audio}}, \boldsymbol{\mathrm{e}}_{\textrm{vision}}), 100\Bigg)\\
		&+\sum_{i=1}^{N_\textrm{neg}}\textrm{MSE}\Bigg(\Big[S\Big(\boldsymbol{\mathrm{e}}^{-}_{\textrm{audio}_i}, \boldsymbol{\mathrm{e}}_{\textrm{vision}}), S(\boldsymbol{\mathrm{e}}_{\textrm{audio}}, \boldsymbol{\mathrm{e}}^-_{\textrm{vision}_i}), S(\boldsymbol{\mathrm{e}}_{\textrm{audio}}, \boldsymbol{\mathrm{e}}^-_{\textrm{vision background}_i})\Big], 0\Bigg)\\
		&+ \sum_{i=1}^{N_\textrm{pos}}\textrm{MSE}\Bigg(\Big[S(\boldsymbol{\mathrm{e}}_{\textrm{audio}}, \boldsymbol{\mathrm{e}}_{\textrm{vision}_i}^+), S(\boldsymbol{\mathrm{e}}_{\textrm{audio}_i}^+, \boldsymbol{\mathrm{e}}_{\textrm{vision}})\Big], 100\Bigg),
		\label{eq:loss}
\end{split}\end{equation}}
where $S$ is calculated with the word-to-image attention mechanism described above. 
Intuitively this should push $\boldsymbol{\mathrm{e}}_{\textrm{audio}}$, $\boldsymbol{\mathrm{e}}_{\textrm{vision}}$ and the positive examples in $\boldsymbol{\mathrm{e}}_{\textrm{audio}_{1:N_{\textrm{pos}}}}^+$ and $\boldsymbol{\mathrm{e}}_{\textrm{vision}_{1:N_{\textrm{pos}}}}^+$ closer. 
At the same time it should push the negative examples in $\boldsymbol{\mathrm{e}}^{-}_{\textrm{audio}_{1:N_{\textrm{neg}}}}$, $\boldsymbol{\mathrm{e}}^{-}_{\textrm{vision}_{1:N_{\textrm{neg}}}}$  and $\boldsymbol{\mathrm{e}}^{-}_{\textrm{vision background}_{1:N_{\textrm{neg}}}}$ away from these positives.
I.e.\ it should learn the visual depiction of a spoken word class.
Therefore, we need positive ($\boldsymbol{\mathrm{e}}_{\textrm{audio}_{1:N_{\textrm{pos}}}}^+$, $\boldsymbol{\mathrm{e}}_{\textrm{vision}_{1:N_{\textrm{pos}}}}^+$) and negative ($\boldsymbol{\mathrm{e}}^{-}_{\textrm{audio}_{1:N_{\textrm{neg}}}}$, $\boldsymbol{\mathrm{e}}^{-}_{\textrm{vision}_{1:N_{\textrm{neg}}}}$, $\boldsymbol{\mathrm{e}}^{-}_{\textrm{vision background}_{1:N_{\textrm{neg}}}}$) pairs. 

\subsection{Few-shot pair mining}
\label{subsec:mining}
For few-shot training, we only have the small number of ground truth examples 
in the support set $\mathcal{S}$.
This would not be sufficient to train the complex model described above. 
To overcome this, \cite{nortje_direct_2021} proposed a pair mining scheme.
We follow the same high-level idea to mine word-word and image-image pairs: use the audio examples in $\mathcal{S}$ and compare each example to each utterance in a large collection of unlabelled audio utterances, and similarly for the images.
The mined items can then be used to construct more word-image pairs for training.
While in~\cite{nortje_direct_2021} the unlabelled collection of audio consisted of isolated spoken words (which was artificially segmented), here we consider an unlabelled collection of audio consisting of full spoken utterances (a more realistic scenario).

The simple isolated-word comparison approach used in~\cite{nortje_direct_2021} is not adequate for this setting. 
We therefore employ another approach.
We have a 
spoken word in our support set that we want to match to unlabelled unsegmented utterances in a large audio collection. 
This is similar to fuzzy string search, i.e.\ finding a set of strings that approximately match a given pattern.
However, algorithms from string search are not directly applicable to speech since they operate on a discrete alphabet. 
To bridge this gap, we use QbERT (query-by-example with HuBERT). 
The idea is to encode speech as a set of discrete units that approximate phones. 
Then we can apply standard string search algorithms to find examples that match a given query word.
We use HuBERT \cite{hsu_hubert_2021} to map input speech into discrete units. 
Then we divide the units into variable-duration phone-like segments following \cite{kamper_towards_2021}.
Finally, we search the dataset by aligning the query to each utterance using the Needleman-Wunsch algorithm \cite{needleman_general_1970}.
An alternative to \qbert\ would have been to use 
dynamic time warping (DTW), 
as is done in~\cite{nortje_direct_2021}. However, in a developmental experiment we found that DTW achieves an isolated word retrieval $F_1$ score of  76.8\% while \qbert\ achieves 98.7\%.


Using \qbert, we compare each spoken utterance in an unlabelled collection of audio utterances to each spoken word example in $\mathcal{S}$.
For each utterance, we take the highest score across the $K$ word examples per class and rank the utterances from highest to lowest for each class. 
The first $n$ utterances with the highest scores for a class are predicted to contain the spoken form of the word.
Additionally, we use \qbert's predicted word segments to isolate matched words. 
To mine image pairs, we follow the same steps, but instead we use AlexNet~\cite{krizhevsky_imagenet_2017} to extract a single embedding for each image and use cosine distance to compare image embeddings to one another.
To get word-image pairs, we mine an image from the same predicted class as a segmented word.
Negative pairs are taken from the positive pairs of other classes.
We also mine an extra negative image from a set known to not contain any of the few-shot classes. 
Therefore, during the few-shot retrieval task, images containing few-shot classes can be distinguished from images that depicts none of the few-shot classes.


\section{Experimental setup}
\label{sec:setup}

\textbf{Data:} 
For our experiments, we use the SpokenCOCO Corpus~\cite{hsu_text-free_2021} which  consists of the MSCOCO~\cite{lin_microsoft_2014} images with recorded spoken captions corresponding to the MSCOCO textual captions.
We 
parametrise the utterances 
as mel-spectograms with a hop length of 10~ms, a window 
of 25~ms and 40~mel bins. 
These are truncated or zero-padded to 1024 frames. 
Images are resized to 224$\times$224 pixels and normalised with means and variances calculated on ImageNet~\cite{deng_imagenet_2009} with VGG~\cite{simonyan_very_2015}.

We use the SpokenCOCO setup of \cite{miller_t_exploring_2022}
by splitting it into two main sets: a set which contains only the few-shot classes and a set that does not contain any of the few-shot classes (listed below).
We refer to the latter set as background data. 
The set containing the few-shot classes is split into training and testing sets. 
We sample the support set $\mathcal{S}$ from this training set (\S\ref{sec:task}) and use the Montreal forced aligner~\cite{mcauliffe_montreal_2017} to isolate the few-shot words.
For our mining approach (\S\ref{subsec:mining}), we need unlabelled audio and image data to mine pairs from; for this we use the remainder of the training data that does not include the support set.
From these unlabelled collections, we mine pairs: the $n=600$ highest ranking examples per class 
(\S\ref{subsec:mining}).
These pairs are split into training and validation pairs.

\textbf{Models:} Figure~\ref{fig:model} illustrates our model, \model\ 
(\S\ref{subsec:model}).
For the image branch, we use a pretrained adaptation of AlexNet~\cite{krizhevsky_imagenet_2017} to get a sequence of per pixel embeddings. 
We use an adaptation of \cite{nortje_towards_2022}'s audio network for the audio branch pretrained in an self-supervised manner on Libri-Light~\cite{kahn_libri-light_2020} and multilingual (English and Hindi) Places~\cite{harwath_vision_2018}.
The model is initialised by pretraining it on the background data using the contrastive retrieval loss of \cite{harwath_unsupervised_2016}.
We take $N_\textrm{pos}=5$ and $N_\textrm{neg}=11$ in Equation~\ref{eq:loss} after fine-tuning it on the development pairs.
For validation, we use the development set to get one positive image $\boldsymbol{\mathrm{x}}_{\textrm{vision}}^+$ and one negative image $\boldsymbol{\mathrm{x}}_{\textrm{vision}}^-$ for each developmental word-image $(\boldsymbol{\mathrm{x}}_{\textrm{audio}},\, \boldsymbol{\mathrm{x}}_{\textrm{vision}})$ pair. 
The validation task measures whether the model will place $\boldsymbol{\mathrm{x}}_{\textrm{vision}}$ and  $\boldsymbol{\mathrm{x}}_{\textrm{vision}}^+$ closer to  $\boldsymbol{\mathrm{x}}_{\textrm{audio}}$ than it would to  $\boldsymbol{\mathrm{x}}_{\textrm{vision}}^-$. 
We train all models with Adam~\cite{kingma_adam_2015} for 100 epochs using the validation task for early stopping. \footnote{Source code: \href{https://github.com/LeanneNortje/Mulitmodal_few-shot_word_acquisition}{Support set}; \href{https://github.com/LeanneNortje/Indirect_few-shot_multimodal_word_acquisition}{\smallmodel\, no mining}; \smallmodel: \href{https://github.com/LeanneNortje/Multimodal_100-shot_5-way_word_acquisition}{100-shot 5-way}; \href{https://github.com/LeanneNortje/Multimodal_50-shot_5-way_word_acquisition}{50-shot 5-way}; \href{https://github.com/LeanneNortje/Multimodal_10-shot_5-way_word_acquisition}{10-shot 5-way}; \href{https://github.com/LeanneNortje/Multimodal_5-shot_5-way_word_acquisition}{5-shot 5-way};
\href{https://github.com/LeanneNortje/Multimodal_5-shot_40-way_word_acquisition}{5-shot 40-way}}

\textbf{Few-shot tasks:} Using the same five classes -- \textit{broccoli}, \textit{zebra}, \textit{fire hydrant}, \textit{sheep} and \textit{kite} -- as \cite{miller_t_exploring_2022}, we evaluate our approach on two tasks (as explained in \S\ref{sec:task}): a traditional few-shot classification task and a few-shot retrieval task.
For both these tasks, the $K$-shot $L$-way support set $\mathcal{S}$ contains $K$ ground truth spoken word-image pairs for each of the $L=5$ classes and is used to mine pairs for training and development.
For testing the few-shot classification task, we sample $1000$ episodes where each episode contains $L$ spoken word queries $\boldsymbol{\mathrm{x}}^q_{\textrm{audio}}$, one for each class, and a matching set $\mathcal{M}$ which contains one image $\boldsymbol{\mathrm{x}}_{\textrm{vision}}^{(i)}$ for each class.   
However, in the few-shot retrieval task, instead of having one image per class, $\mathcal{M}$ consists of $5000$ images $\boldsymbol{\mathrm{x}}_{\textrm{vision}}^{(i)}$ where some depicts a few-shot class and others do not. 
Here, 20 query words are taken per class and averaged to get $\boldsymbol{\mathrm{x}}^q_{\textrm{audio}}$. 
For each of the $L$ queries $\boldsymbol{\mathrm{x}}^q_{\textrm{audio}}$, these $5000$ images are ranked from highest to lowest similarity.
The precision at $N$ ($P@N$) score is the proportion of images in the top $N$ highest ranking images that are from the same class as $\boldsymbol{\mathrm{x}}^q_{\textrm{audio}}$.
$N$ is the actual number of images in $\mathcal{M}$ that depicts the word class. 

\section{Experimental results}

\begin{table}[tb]
	\caption{
		$P$@N few-shot retrieval scores (\%) on the five few-shot classes. $K$ is the number of support-set examples per class.	}
	\vspace{-5pt}
	\label{tab:previous_work}
	\centering
	\footnotesize
	\setlength{\tabcolsep}{0.25em}
	\renewcommand{\arraystretch}{1.2}
	\begin{tabular}{ lcccc }
		\toprule
		\multicolumn{1}{c}{Model} & 
		\multicolumn{4}{c}{$K$}\\
		\cline{2-5}
		 & 5 & 10 & 50 & 100 \\
		\midrule
		Naive fine-tuned~\cite{miller_t_exploring_2022} & -- & -- & -- & \textbf{52.5}\\
		Oracle masking~\cite{miller_t_exploring_2022} & -- & 8.4$\pm$0.0  & 24.0$\pm$0.1 & 35.5$\pm$0.2\\
		\smallmodel\ & \textbf{44.4$\pm$0.0} & \textbf{43.4$\pm$0.1} & \textbf{40.2$\pm$0.0} & 42.5$\pm$0.1\\
		\addlinespace
		\smallmodel, no mining\hspace{0.5em} & 22.0$\pm$0.4 & 24.1$\pm$0.8 & 22.7$\pm$0.5 & 23.2$\pm$1.1\\
		\smallmodel, $L_{\textrm{train}}=40$\hspace{0.5em} & 39.7$\pm$0.6 & -- & -- & --\\
		\bottomrule
	\end{tabular}
\end{table}

We first want to compare our work directly to that of~\cite{miller_t_exploring_2022}. Concretely, we compare to two of \cite{miller_t_exploring_2022}'s models on the few-shot retrieval task.
The first model we compare to is their naive model, which is trained on background classes and fine-tuned on $K=100$ examples for each of the $L=5$ classes.
The second is their oracle masking model in which the contrastive loss used during fine-tuning 
ensures that a negative image does not contain any instance of the anchor few-shot class.
The results are given in Table~\ref{tab:previous_work} (\cite{miller_t_exploring_2022} did not report scores for fewer than $K = 10$).

Comparing our full \model\ model to the oracle masking approach, we see that we outperform  \cite{miller_t_exploring_2022} consistently across all values of $K$. 
Neither \model\ or oracle masking works as well as the naive fine-tuned approach (line 1), but fine-tuning only works with a large number of shots.
We also see that our approach (a bit surprisingly) delivers approximately the same few-shot retrieval scores as $K$ increases, whereas \cite{miller_t_exploring_2022}'s scores in line~2, increase.
The reason for this is that the models retain contextual information which makes it difficult to disentangle the images containing a few-shot class from background images.
However, our approach works particularly well with fewer shots.

To determine the mined pairs' contribution to this performance boost, we do an experiment where we do not update \model\ on the mined pairs after pretraining it on the background data (not containing any of the few-shot classes). 
To test this model on the few-shot retrieval task, we use the indirect few-shot method of \cite{eloff_multimodal_2019, nortje_unsupervised_2020}: each $\boldsymbol{\mathrm{x}}^q_{\textrm{audio}}$ is compared to each $\boldsymbol{\mathrm{x}}_{\textrm{audio}}^{(j)}$ in $\mathcal{S}$ to find the audio example closest to the query.
The image $\boldsymbol{\mathrm{x}}_{\textrm{vision}}^{(j)}$ corresponding to the closest $\boldsymbol{\mathrm{x}}_{\textrm{audio}}^{(j)}$ is then used to calculate the similarity to each image $\boldsymbol{\mathrm{x}}_{\textrm{vision}}^{(i)}$ in $\mathcal{M}$.
Using mined pairs improves the scores with roughly 20\% when comparing lines~3 and 4.
In the final line of Table~\ref{tab:previous_work} we start to investigate how our approach performs when, instead of using just five few-shot classes, we have 40 few-shot classes to learn with $K = 5$ shots. We see that we pay roughly 5\% in $P@N$, but even when learning 40 classes, we still outperform the five-class oracle masking approach across all shots considered.

\begin{table}[tb]
	\caption{
		Few-shot word classification accuracy scores (\%).
		We vary the number of shots per class $K$. 
		Instead of only considering the five classes from~\cite{miller_t_exploring_2022}, we also look at settings with 40 classes in the support and/or matching sets.}
	\vspace{-5pt}
	\label{tab:few-shot}
	\centering
	\footnotesize
	\renewcommand{\arraystretch}{1.2}	
	\begin{tabular}{ cccccc}
		\toprule
		\multicolumn{1}{c}{Model}&
		\multicolumn{1}{c}{$K$}&
		\multicolumn{1}{c}{$L_{\mathcal{S}}$}&
		\multicolumn{1}{c}{$L_{\mathcal{M}}$}&
		\multicolumn{1}{c}{Few-shot accuracy}\\
		\midrule
		\multirow{4}{2cm}{\centering\smallmodel, no mining} & 5 & -- & 5 & 50.4\\
		& 10 & -- & 5 & 48.0\\
		& 50 & -- & 5 & 48.5\\
		& 100 & -- & 5 & 47.7\\
		\hline
		\multirow{7}{2cm}{\centering\smallmodel, with mining} & 5 & 5 & 5 & 65.4\\
		& 10 & 5 & 5 & 77.5\\
		& 50 & 5 & 5 & 86.6\\
		& 100 & 5 & 5 & \textbf{90.9}\\
		\addlinespace
		& 5 & 40 & 5 & 63.7\\
		& 5 & 40 & 40 & 21.7\\
		\bottomrule
	\end{tabular}
\end{table}

To further analyse the performance gains from mining and to also see what happens with more classes, we now consider the conventional few-shot word classification task (\S\ref{sec:task}). 
This task wasn't used in~\cite{miller_t_exploring_2022}.
Table~\ref*{tab:few-shot} shows that the few-shot classification scores increase as $K$ increases when we use mined pairs.
For the no pair mining method, the scores are worse and drops slightly as $K$ increases.
Looking at the five-shot case, we see that training on more classes ($L_{\mathcal{S}}=40$) leads to a slightly lower classification score on the original five classes.
The few-shot accuracy when tested on all 40 classes is 21.7\%. 

All together, our results sets a competitive multimodal baseline for both few-shot retrieval and word classification in settings where the number of shots is small.

\section{Conclusion}

Our goal was to do multimodal few-shot learning of natural images and spoken words. 
To do this, we proposed a novel few-shot pair mining method which we use in a new multimodal word-to-image attention model. 
For the lower-resource scenario where $K$ is small, our model achieves higher few-shot retrieval scores than an existing model.
We also set a competitive baseline for natural visually grounded few-shot word classification and present preliminary experiments indicating that our approach can be used on more than five few-shot classes.
Future work will look into improving few-shot word classification on more~classes.

\section{Acknowledgements}

We would like to thank DeepMind for funding Leanne Nortje and Google for funding Benjamin van Niekerk.
We would also like to thank Tyler Miller and David Harwath for helping with the few-shot retrieval comparisons.

\bibliographystyle{IEEEtran}

\bibliography{mybib.bib}


\end{document}